\documentclass{ecai} 
\usepackage{multirow}
\usepackage{latexsym}
\usepackage{amssymb}
\usepackage{amsmath}
\usepackage{amsthm}
\usepackage{booktabs}
\usepackage{enumitem}
\usepackage{graphicx}
\usepackage{color}
\usepackage{utfsym}
\usepackage{makecell}
\usepackage{float}
\usepackage{stfloats}
\usepackage{url}
\usepackage[hidelinks]{hyperref}
\newcommand{\BibTeX}{B\kern-.05em{\sc i\kern-.025em b}\kern-.08em\TeX}

\begin{document}

%%%%%%%%%%%%%%%%%%%%%%%%%%%%%%%%%%%%%%%%%%%%%%%%%%%%%%%%%%%%%%%%%%%%%%%%

\begin{frontmatter}

%%% Use this command to specify your submission number.
%%% In doubleblind mode, it will be printed on the first page.

\paperid{843} 

%%% Use this command to specify the title of your paper.

\title{First Creating Backgrounds Then Rendering Texts: \\A New Paradigm for Visual Text Blending}

%%% Use this combinations of commands to specify all authors of your 
%%% paper. Use \fnms{} and \snm{} to indicate everyone's first names 
%%% and surname. This will help the publisher with indexing the 
%%% proceedings. Please use a reasonable approximation in case your 
%%% name does not neatly split into "first names" and "surname".
%%% Specifying your ORCID digital identifier is optional. 
%%% Use the \thanks{} command to indicate one or more corresponding 
%%% authors and their email address(es). If so desired, you can specify
%%% author contributions using the \footnote{} command.

\author[A,C]{\fnms{Zhenhang}~\snm{Li}}
\author[A]{\fnms{Yan}~\snm{Shu}}
\author[A,C]{\fnms{Weichao}~\snm{Zeng}}
\author[A,C]{\fnms{Dongbao}~\snm{Yang}}
\author[B]{\fnms{Yu}~\snm{Zhou}\thanks{Corresponding Author. Email: yzhou@nankai.edu.cn.}}

\address[A]{Institute of Information Engineering, Chinese Academy of Sciences, Beijing, China}
\address[B]{TMCC, College of Computer Science, Nankai University, Tianjin, China}
\address[C]{School of Cyber Security, University of Chinese Academy of Sciences, Beijing, China}

%%% Use this environment to include an abstract of your paper.

\begin{abstract}
Diffusion models, known for their impressive image generation abilities, have played a pivotal role in the rise of visual text generation. Nevertheless, existing visual text generation methods often focus on generating entire images with text prompts, leading to imprecise control and limited practicality. A more promising direction is visual text blending, which focuses on seamlessly merging texts onto text-free backgrounds. However, existing visual text blending methods often struggle to generate high-fidelity and diverse images due to a shortage of backgrounds for synthesis and limited generalization capabilities. To overcome these challenges, we propose a new visual text blending paradigm including both creating backgrounds and rendering texts. Specifically, a background generator is developed to produce high-fidelity and text-free natural images. Moreover, a text renderer named GlyphOnly is designed for achieving visually plausible text-background integration. GlyphOnly, built on a Stable Diffusion framework, utilizes glyphs and backgrounds as conditions for accurate rendering and consistency control, as well as equipped with an adaptive text block exploration strategy for small-scale text rendering. We also explore several downstream applications based on our method, including scene text dataset synthesis for boosting scene text detectors, as well as text image customization and editing. Code and model will be available at \url{https://github.com/Zhenhang-Li/GlyphOnly}.
\end{abstract}

\end{frontmatter}

%%%%%%%%%%%%%%%%%%%%%%%%%%%%%%%%%%%%%%%%%%%%%%%%%%%%%%%%%%%%%%%%%%%%%%%%

\section{Introduction}
In recent years, diffusion models~\citep{ddpm} have made considerable advancements in image generation. The emergence of Latent Diffusion Models~\cite{latent-diffusion} has enabled a breakthrough in text-to-image generation. Yet, producing legible and high-fidelity visual texts is still a challenging task~\cite{shu2024visual}, owing to the complex nature of texts, such as diverse fonts, varied styles, and intricate glyph details. To address these challenges, numerous methods have been introduced, focusing on enhancing the conditional text encoder \cite{deepfloyd,Imagen} or incorporating glyph guidance \cite{glyphdraw,anytext,glyphcontrol} for precise rendering.

\begin{figure}[ht]
\begin{center}
    
\includegraphics[width=1.0\linewidth]{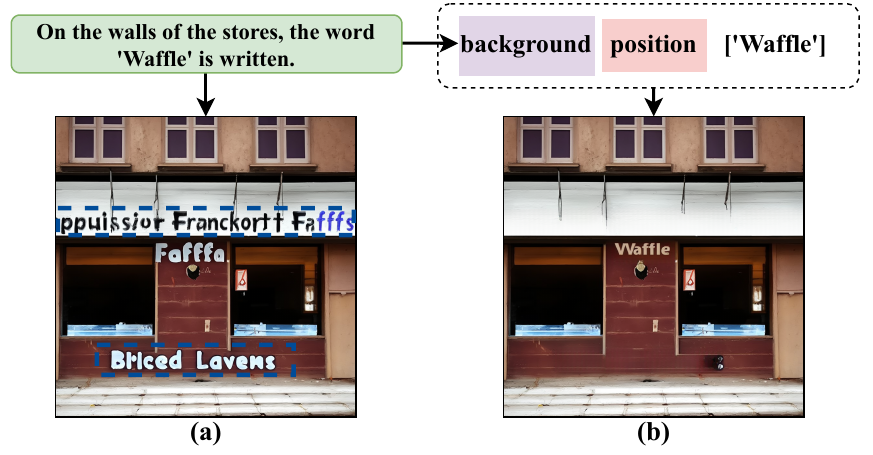}
    \caption{Visual Texts generated by (a) Existing Visual Text Generation Methods, and (b) Our Visual Blending Method.}
\label{fig:first}
\end{center}
\end{figure}

However, most visual text generation methods focus on producing an entire image based on a text prompt, which leads to two limitations: (i) \textbf{Imprecise control over the generated texts, including their quantity and layout.} Due to the inherent characteristics of conditional encoders, users are unable to generate a large volume of texts in complex layouts. (ii) \textbf{Inflexible control over the generated backgrounds.} Users are unable to render text on a specific background, nor can they guarantee the generated scene does not include unintended textual elements, as shown in Figure~\ref{fig:first} (a). Consequently, the practical applications of these methods are somewhat restricted.

In this paper, we shift the focus to visual text blending, which aims to mix texts on specified backgrounds seamlessly. This field has a long history of research. Studies~\cite{SynthText,synthtext3d,unrealtext,visd} adopt an image composition approach, which aims to optimize surface smoothness in the combined images. Based on the Generative Adversarial Networks (GANs), learning-based methods \cite{stsgan,sfgan} have been employed to replicate the realistic appearance of actual text using reference samples. However, these methods lack robustness and often fail to generate images with high fidelity and diversity, primarily for two reasons: (i) There is a shortage of sufficiently diverse backgrounds for synthesis and training a robust visual text renderer; (ii) The models used for text rendering exhibit limited generalization capabilities, struggling with rendering texts in various styles. More recently, diffusion-based methods~\cite{textdiffuser,diffste} have adopted an inpainting framework for text rendering. These approaches show limited accuracy in text rendering and visual consistency, particularly when there is insufficient surrounding visual text to reference. Additionally, these methods struggle with rendering text in small sizes and complex arrangements.

To address these challenges, we propose a new visual text blending paradigm - first creating backgrounds then rendering texts, as shown in Figure~\ref{fig:first} (b). Specifically, we design a background generator that integrates existing expert models to produce high-quality, text-free background images. Obtaining text-free backgrounds is essential for generating diverse visual text images that can be utilized in downstream tasks~\cite{lyu2024textblockv2,shu2023perceiving,yang2024masked}. Furthermore, we develop GlyphOnly, tailored specifically for visual text rendering. GlyphOnly stands out from most existing diffusion-based visual text generation methods in that it relies solely on glyph images as conditions, rather than natural language prompts. To enhance visual consistency and text rendering accuracy, we incorporate prior background features into the condition encoder and introduce text sequence recognition supervision during the denoising process. To render small-scale texts due to limitations inherent in Variable Autoencoders (VAEs), we propose an adaptive text block exploration strategy without increasing computational complexity.

We also explore various downstream applications leveraging our proposed visual text blending paradigm. For instance, we have created a synthetic scene text dataset, termed SynthGlyph, using our proposed semantic-aware position selection algorithm. SynthGlyph notably enhances the efficacy of current scene text detector. Furthermore, our method is also applicable for text image customization and editing.

We summarize the contributions of our method as follows:

\begin{itemize}
\item To the best of our knowledge, this is the first work to focus on improving the quality and diversity of backgrounds for visual text blending. A new paradigm from creating diverse backgrounds to rendering various texts is proposed.  

\item We integrate existing expert models to design a text-free background generator, which facilitates the training of a robust visual text renderer and ensures the generated images with high diversity. 

\item A diffusion-based model designed for visual text rendering, GlyphOnly, is proposed. Beyond achieving high text rendering accuracy and exceptional visual realism, GlyphOnly is adept at rendering small-scale texts legibly.

\item We explore various downstream tasks utilizing our proposed paradigm. Experiments have proven that our synthetic data can boost the performance of existing scene text detectors noticeably. Besides, our work demonstrates potential in other applications like text image customization and editing.  
\end{itemize}

\section{Related Work}

\subsection{Visual Text Generation}
The advancement of Diffusion Models \cite{ddpm,latent-diffusion} has led to a plethora of methods for creating high-quality images. Yet, producing legible and visually coherent texts remains a challenge. To address this, Imagen \cite{Imagen} and DeepFolyd \cite{deepfloyd} employ the large-scale language model T5 to enhance text spelling comprehension. Research by \cite{Character-aware} indicates that character-aware models like ByT5 \cite{byt5} have distinct advantages over character-blind models such as T5 and CLIP. GlyphDraw introduces a unique framework for precise character generation control, incorporating auxiliary text locations and glyph features. TextDiffuser combines a Layout Transformer \cite{layout} to acquire text arrangement knowledge, along with character-level segmentation masks for better text rendering precision. GlyphControl adopts a ControlNet-based framework \cite{Controlnet} that facilitates explicit learning of text glyph features. Diff-Text~\cite{brushyourtext} leverages rendered sketch images as priors, thus arousing the potential multilingual-generation ability of the pre-trained Stable Diffusion. 

While the aforementioned methods have yielded promising results, they are relatively inflexible to control backgrounds and generated texts. This is because they primarily focus on generating entire images based on text prompts, rather than seamlessly blending specific texts onto designated backgrounds.

\subsection{Visual Text Blending}
Visual text blending methods, aimed at addressing the lack of visual coherence resulting from simple text overlay on backgrounds, have undergone extensive exploration. SynthText \cite{SynthText} identifies suitable text placement areas using depth and segmentation maps, then embeds texts via perspective transformation. VISD \cite{visd} employs semantic segmentation to pinpoint optimal text generation regions and enhances visual quality by choosing more fitting text colors. SynthText3D \cite{synthtext3d} and UnrealText \cite{unrealtext} produce text images within 3D scenes using game engines, thereby heightening the realism of the generated images. In the realm of GANs, SF-GAN \cite{sfgan} and STS-GAN \cite{stsgan} have been developed to learn the blending modes, including geometric and appearance aspects, between texts and real image backgrounds.

Current methods often face challenges in rendering texts accurately and achieving visual coherence with the surroundings, primarily due to a deficiency of diverse backgrounds for training and synthesis.

In this paper, we produce a new paradigm in visual text blending, first creating backgrounds then rendering texts to mitigate these issues.

\section{Method}
\begin{figure*}[ht]
    \begin{center}    
\includegraphics[width=1.0\linewidth]{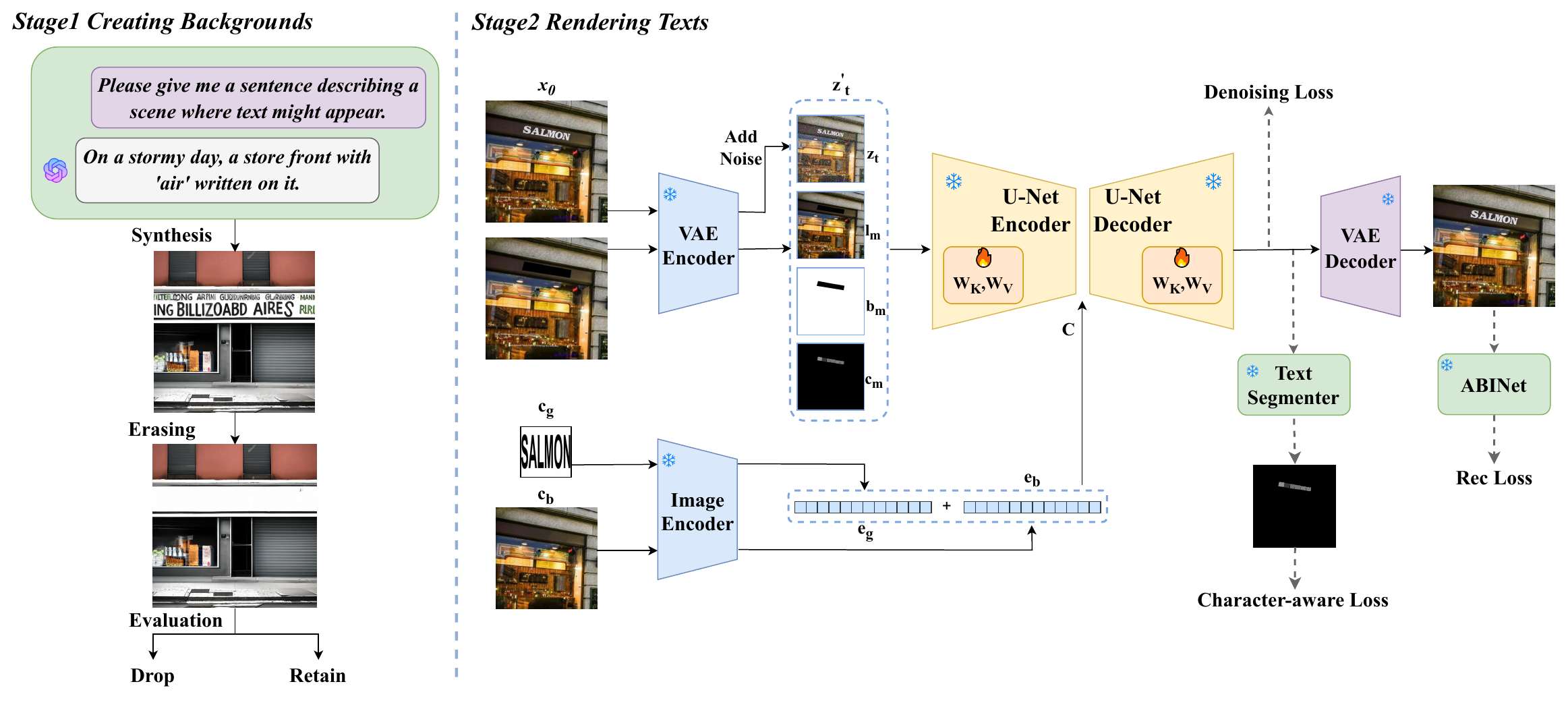}
    \vspace{-20pt}
    \caption{The framework of the proposed method. The first stage is creating background, which involves synthesis, erasing and evaluation. In the second rendering texts stage, GlyphOnly integrates noisy features, segmentation masks, feature masks, and masked features as inputs to the U-Net. The frozen pre-trained CLIP Image Encoder converts glyph images and background images into embeddings for generation control. During training, only the parameters of the convolutional layers of the U-Net input, the convolutional layers of the conditional input, and the key and value components of the U-Net cross-attention layers are updated. Please be aware that the diffusion model performs denoising in the latent space, but we utilize image pixels for better visualization. 
    }
\label{fig:pipeline}
\end{center}
\vspace{-5pt}
\end{figure*}
\subsection{Creating Backgrounds}

Securing text-free backgrounds that closely resemble the natural distribution of real-world images is crucial for synthesizing diverse visual text images that can be applied to downstream tasks. However, this task presents notable challenges due to several factors: (i) Directly capturing natural images in the real world is limited in quantity and labor-intensive. Moreover, it is challenging to meet customized requirements; (ii) Existing text-to-image models are unable to synthesize text-free images, even with carefully crafted prompts; (iii) It is a difficult issue to remove existing texts in generated images; 

Echoing the adage ``Standing on the shoulders of giants", we utilize pre-existing expert models and integrate them into a powerful background generator. This enables the synthesis of limitless text-free, high-fidelity images, as illustrated in Figure~\ref{fig:pipeline} (Stage 1). The whole process can be divided into three steps, namely synthesis, erasing, and evaluation. 

\paragraph{Synthesis} We utilize the openly available text-to-image model, DeepFloyd \cite{deepfloyd}, to synthesize natural images. Furthermore, to generate a large number of images that closely resemble real-world text images, we engage with ChatGPT-3.5-turbo~\cite{chatgpt} for automated prompt design. For instance, when requesting suggestions for scene text image generation, ChatGPT provides numerous responses, such as ``On a stormy day, a store front with `air' written on it." Using this method, we can efficiently gather large-scale natural images in batches.

\paragraph{Erasing} To acquire clean images devoid of text, we use a pre-trained inpainting model \cite{suvorov2022LaMa} to erase the texts. For this process, we employ an available OCR API \cite{ocr} to provide text region masks, thereby aiding the text removal.

\paragraph{Evaluation} To ensure high image quality, we conduct evaluations to filter out low-fidelity images. Specifically, we consider two aspects: (i) PSNR to assess overall visual quality, as determined by a non-reference image quality method \cite{quality}; (ii) Text residual evaluation using \cite{ocr}.

\subsection{Rendering Texts}
\paragraph{Latent Diffusion Models} LDMs~\cite{latent-diffusion} are newly introduced variants of Diffusion Models. Compared to the DDPMs \cite{ddpm} that operate in pixel space, LDMs perform denoising process in latent space. They first utilize a pre-trained autoencoder $E$ to compress images $x$ into latent representations $z_0=\varepsilon(x)$, and apply a decoder $D$ to reconstruct the latent back to pixel space, such that $D(E(x)) \approx x$. Based on the cross-attention mechanism, various conditions $C$ can be integrated into the framework, which has following objectives:

\begin{equation}
\label{eq1}
    \mathcal{L}_{denoising} = \mathbb{E}_{\epsilon(x_0),C,\epsilon \sim \mathcal N(0,1),t} \big[\Vert \epsilon - \epsilon_\theta(z_t, t, C)\Vert^2_2\big].
\end{equation}

Here, $z_t$ is the perturbed $z_0$ and $\epsilon_\theta$ is implemented by a conditional U-Net \cite{unet}  model.

\paragraph{Inpainting Architecture} Built on the foundation of LDMs, we propose an inpainting architecture dubbed GlyphOnly, in order to realize high-realism visual text rendering. The overview of GlyphOnly is illustrated
in Figure~\ref{fig:pipeline} (Stage 2).

Inspired by TextDiffuser \cite{textdiffuser}, we enhance the base LDMs with auxiliary guidance, including unfilled image references and text positions.  This guidance comprises  $l_m$, the latent vector of the image with masked text regions; $b_m$, the word-level mask; and $c_m$, the character segmentation mask. To address expressiveness and channel alignment with the VAE, we introduce a stack of convolutional layers to decode the feature and inject it into the diffusion process. The $z_t$ in Eq.~\ref{eq1} is redefined as $z_t'$ :

\begin{equation}
  z_t'=Conv(z_t \oplus l_m \oplus b_m \oplus c_m),
\end{equation}
where $\oplus$ denotes the concatenation operation along the channel dimension.

\paragraph{Conditions Designing}
Typically, conditions $C$ are embeddings encoded from text prompts. To bridge the substantial domain gap and effectively achieve our visual text blending objective, we substitute the natural language representation with glyph image $c_g$, thereby enabling more accurate text rendering. Additionally, given the limited background visual prompts from $l_m$, particularly when employing the adaptive text block exploration strategy (discussed in the subsequent section), we incorporate background images $c_b$ obtained in stage 1 into the conditions. Both $c_g$ and $c_b$ are encoded by a pre-trained CLIP image encoder to obtain the embeddings $e_g$ and $e_b$. The formal conditions can be defined as:          

\begin{equation}
  C=Conv(e_g \oplus e_b).
\end{equation}

\paragraph{Loss Functions} Following \cite{textdiffuser}, we segment the latent features to obtain character-level segmentation masks by utilizing a pre-trained character segmentation model, and we use cross-entropy loss $\mathcal{L}_{char}$ as the character-aware loss.

To enhance the text rendering accuracy, we take the text sequence features as consideration. To this end, we crop the interested text regions from the decoded image, calculating the text recognition loss $\mathcal{L}_{rec}$ by a pre-trained ABINet~\cite{abinet}, which is a cross-entropy loss. Therefore, our total loss function can be expressed as follows:
\begin{equation}
    \mathcal{L}_{total} = \mathcal{L}_{denoising} + \lambda_{char}*\mathcal{L}_{char}+ 
    \lambda_{rec}*\mathcal{L}_{rec}.
\end{equation}

Here, $\lambda_{char}$ and $\lambda_{rec}$ are set to 0.01 and 0.03 respectively.

\begin{table*}[bp]
\begin{center}
    
\caption{Quantitative comparison with existing methods. The bold numbers indicate the highest-performing result.}
\vspace{+10pt}
\begin{tabular}{@{}cccccc@{}}
\toprule
Metrics & \thead{GlyphOnly\\(ours)}  & DiffSTE & TextDiffuser  & AnyText & \thead{GlyphControl\\(direct generation)} \\ \midrule
Accuracy &  \textbf{66.99} & 38.16 & 28.98 & 22.91 &48.00\\ 
1-NED &  \textbf{72.75} & 54.03 & 42.45 & 38.99 & ---\\
\bottomrule
\end{tabular}
\label{tab_reg}   
\end{center}
\end{table*}

\paragraph{Adaptive Text Block Exploration Strategy} While existing diffusion-based methods show promising results in text rendering, they typically face challenges in rendering small-scale texts. Our key observations suggest that this issue is partially due to inherent limitations of VAE in LDMs. Dimensional compression in VAE effectively reduces computational complexity but at the cost of losing some fine-grained texture and glyph features. A direct solution would be to employ an another VAE that generates higher-resolution feature maps, but this approach inevitably increases computational complexity. An alternative method could involve using the text region as input. However, this often results in visual inconsistency, primarily because of the limited information available in $l_m$, where most pixels are masked.

To resolve this dilemma, we introduce the concept of the text block, which is used as input in the diffusion process. It is a balanced approach that provides both a high-resolution representation of fine-grained glyph features and sufficient background pixel priors. Due to the absence of block-level annotations in the dataset, we have developed an adaptive text block exploration strategy.

In our approach, starting with the entire image and a quadrilateral bounding box indicating the text region, we first identify its corresponding minimum enclosing rectangle $R$
with width $w$ and height $h$. Using the centroid of $R$ as a reference, we transform $R$ into an expanded square region $SR$ with side length $s$, guided by the following heuristic rules:

\begin{equation}
s= {2}^{1+\lfloor\log_{2}{\max(w,h)}\rfloor+\lfloor\log_{2}{\lceil\max(w,h)/64\rceil}\rfloor}.
\end{equation}

Then, we crop $SR$ from the original images and resize it to $512 \times 512$ by the bilinear interpolation algorithm. To mitigate the background incompleteness in some cases,  we provide intact background references as a complement to $l_m$.  

\paragraph{Inference Stage} Our inference process has two settings. The first setting is generating background and then rendering text. If we need to generate a large number of backgrounds for downstream tasks, such as text detection datasets, we choose to involve ChatGPT to generate a large number of prompts for background generation. If we only want to obtain a custom image, a prompt can be provided manually. Then we use DeepFloyd to generate the background and obtain a background image without text by using an erasure model. The second setting is direct text rendering on an already available background image. During the inference stage, the manual intervention for position selection in our method is optional. When generating large amounts of data for downstream tasks, we employed automatic position selection, which is described in Section~\ref{section 3.3}. For the character segmentation mask $c_m$, the glyph image is transformed with perspective transformation to fit into a given quadrilateral position. Then, we employ a pre-trained segmentation model to obtain the segmentation mask.

\subsection{Applications}
\label{section 3.3}
\paragraph{Dataset Synthesis for Scene Text Detection} Existing scene text detection methods~\cite{qin2021mask,qin2023towards,tpsnet} require a large quantity of training data. However, acquiring sufficient scene text images and their accurate annotations is labor-intensive and time-consuming. To this end, we utilize our visual text blending paradigm to generate a synthetic scene dataset with annotations for pre-training text detectors.  

To enhance the distribution consistency between synthetic images and real data, we specifically propose a semantic-aware position selection algorithm for automated text rendering region selection. For any given background image, the segmentation map and depth information are obtained using panoramic segmentation~\cite{ODISE} and depth estimation~\cite{zoedepth} techniques, respectively. Subsequently, we identify reasonable regions for text rendering, focusing on pre-defined categories such as ``walls'' and ``signs''. Finally, we follow \cite{SynthText} to refine the selection of rendering regions utilizing semantic and depth data.

\paragraph{Text Image Customization and Editing} Personalizing images with specific texts play a crucial role in various practical applications, including augmented reality and digital marketing. To accomplish this objective, users have the option to select their desired background or create one using our background generator, which accepts text descriptions as input. Subsequently, texts of any size can be realistically rendered at specified positions.

\begin{figure*}[t]
    \begin{center}
\includegraphics[width=0.83\linewidth]{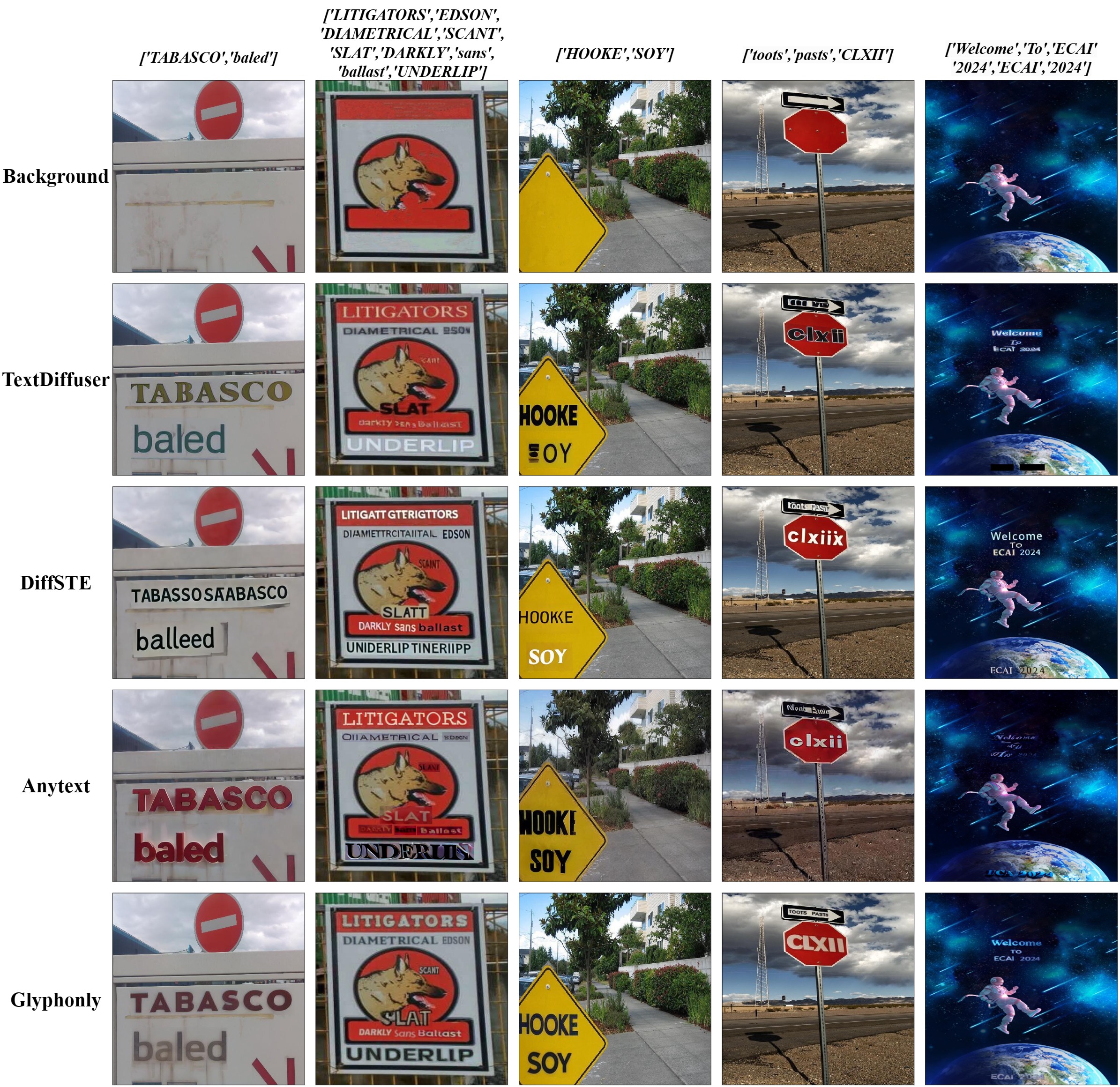}

    \caption{Visualization comparison between our approach and existing methods.}

\label{fig:bijiao1}
\end{center}
\end{figure*}

\begin{figure*}[t]
    \begin{center}
\includegraphics[width=0.83\linewidth]{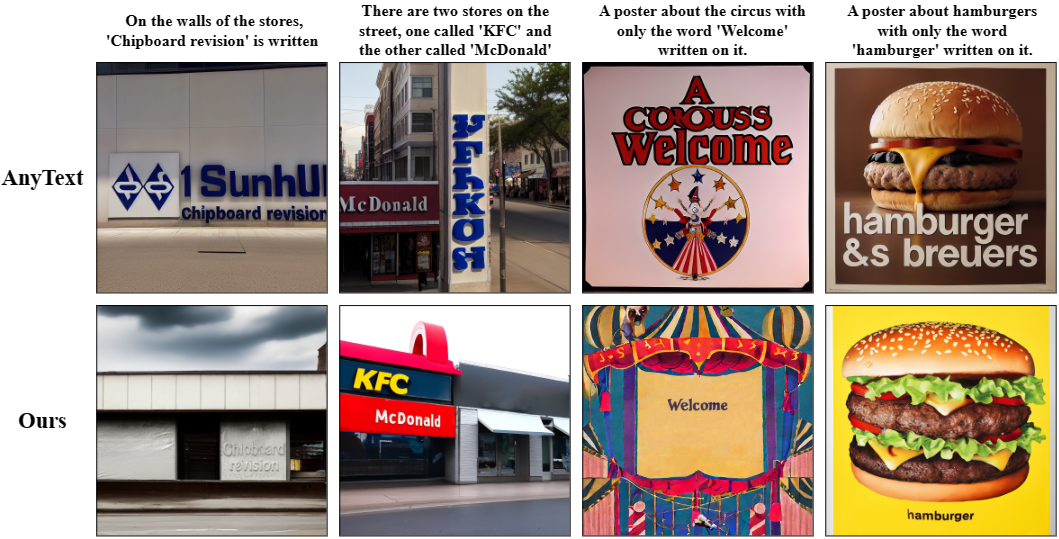}

    \caption{Qualitative comparison results. We compare our method with the SOTA direct generation approach.}
\label{fig:sota}
\vspace{-5pt}
\end{center}

\end{figure*}

\section{Experiments}
\subsection{Datasets}\label{sec:datasets}
To train GlyphOnly model, we utilize several public real scene text datasets. The real data includes the training set from  ICDAR2013 (IC13)~\cite{icdar13}, ICDAR2015 (IC15)~\cite{icdar15}, MLT17~\cite{mlt17}, MLT19~\cite{mlt19}, SCUT-EnsText~\cite{scutenstext}. The total volume of training data amounts to about 60k. We explain the data processing in the Appendix~\cite{li}.

To evaluate the performance of GlyphOnly, we randomly select 500 images with 2,733 text regions from the SCUT-EnsText test set. For each image, a word is randomly chosen from a dictionary containing 88,172 words, which is then used as the text to be generated within the erased region. This methodology enables us to create a specialized test dataset, aimed at assessing the visual text blending capabilities of our model.

\subsection{Implementation Details}
\paragraph{Training of GlyphOnly} We initialize the model with parameters from Stable-Diffusion-v1-5, and employ the parameters from CLIP's image encoder for initializing our image encoder. 

We set the batch size to 32 and train the model for 60 epochs. We utilize the AdamW optimizer and set the learning rate to 1e-5. More details can be seen in the Appendix~\cite{li}. 

\paragraph{Scene Text Detection} As a crucial application, we generate a synthetic dataset for pre-training scene text detectors (See Sec~\ref{sec:Scene Text Detectors}). We adopt DBNet~\cite{dbnet} as our detector with training from scratch. We utilize Stochastic Gradient Descent (SGD) as the optimizer, employing a learning rate of 0.007, a momentum of 0.9, and a weight decay of 1e-4 for training 100,000 iterations. During the fine-tuning stage, we train for 1200 epochs. 

All experiments are implemented in Pytorch on NVIDIA RTX 4090 GPUs.

\begin{table*}
\centering
\vspace{-5pt}
\caption{Scene text detection results of DBNet models pre-trained solely on each synthetic dataset, and tested on real text dataset without fine-tuning.}
\vspace{+10pt}
\begin{tabular}{@{}cccccccccc@{}}
\toprule
\multirow{2}{*}{Training Data} & \multicolumn{3}{c}{IC13} & \multicolumn{3}{c}{IC15} & \multicolumn{3}{c}{TotalText} \\
\cmidrule(l){2-4} \cmidrule(l){5-7} \cmidrule(l){8-10} 
& Precision & Recall & Hmean & Precision & Recall & Hmean  & Precision & Recall & Hmean\\
\midrule
SynthText 10K~ & 75.24 & 63.56 & 68.91 & 70.27 & 46.89 & 56.25  & 60.13 & 53.05 & 56.37\\
VISD 10K & 82.85 & 68.40 & 74.94 & 68.71 & 58.69 & 63.31 & 71.30 & 53.72 & 61.28\\
SynthText3D 10K & 79.58 & 65.11 & 71.62 & 74.08 & 50.36 & 59.96 & 72.08 & 51.87 & 60.33\\
UnrealText 10K & 80.12 & 64.19 & 70.03 & 72.29 & 51.37 & 60.06 & 71.56 & 51.69 & 60.03\\
SynthGlyph 10K & 83.26 & 68.58 & \textbf{75.21} & 70.68 & 59.89 & \textbf{64.84} & 66.52 & 59.82 & \textbf{62.99}\\
\bottomrule
\end{tabular}

\label{tab_pretrain_db}
\end{table*}

\begin{figure}
    \begin{center}
\includegraphics[width=0.8\linewidth]
    {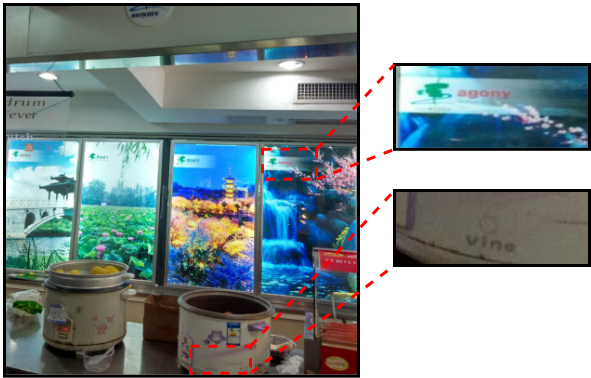}
    
    \caption{Visualization of tiny-size text generation.}
    \vspace{+5pt}
\label{fig:small}
\end{center}

\end{figure}

\begin{figure*}[t]
    \begin{center}
    \includegraphics[width=0.75\linewidth]{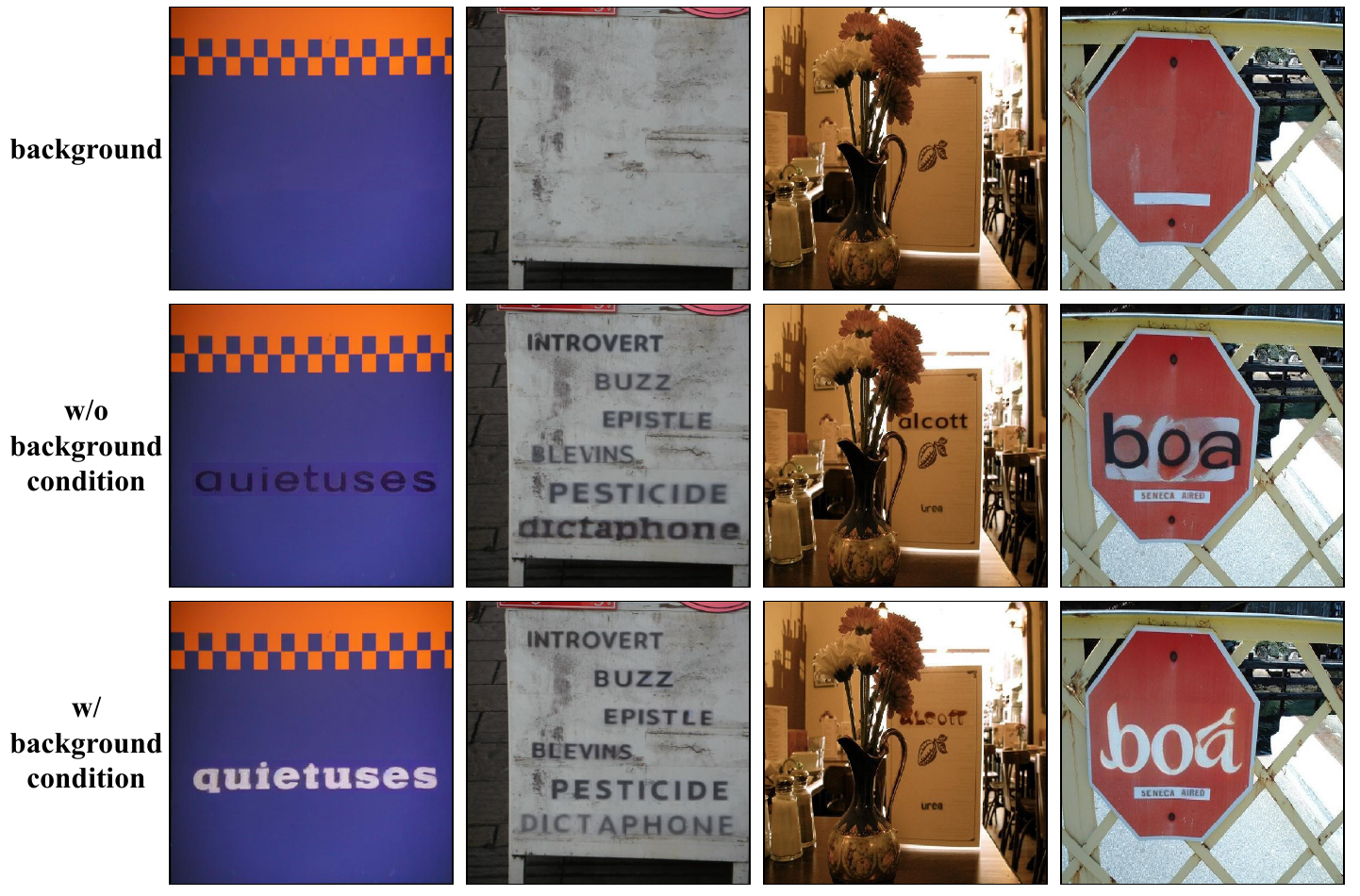}
    \vspace{-5pt}
    \caption{Visualization of the generated text regions with/without background conditions.
    }
\label{fig:bg_ablation}
\end{center}
\vspace{-5pt}
\end{figure*}

\subsection{Comparison with Previous Methods}
\paragraph{Quantitative Comparison} To validate the superiority of our proposed method, we compare it with three recent Diffusion-based visual text blending methods, including TextDiffuser \cite{textdiffuser}, DiffSTE \cite{diffste} and AnyText \cite{anytext}. Additionally, we present experimental results on GlyphControl, which cannot perform text blending and only has generation ability. Through extracting regions of generated texts, we select the following two metrics for comparing text rendering accuracy in word-level and character-level respectively: (1) Text recognition accuracy; (2) Normalized edit distance. As shown in Table \ref{tab_reg}, our method outperforms existing methods by a significant margin in terms of both text recognition accuracy and normalized edit distance. It is worth noting that the datasets we used are only a subset of the AnyText training datasets while achieving higher performance in terms of text blending, which is sufficient to demonstrate the effectiveness of our method. Additionally, please be aware that the benchmark in GlyphControl does not involve generating small-sized text or ensuring the absence of unintended texts.

\paragraph{Qualitative Comparison}  Figure~\ref{fig:bijiao1} illustrates a comparison between our method and existing methods, demonstrating how texts are seamlessly blended into specified regions across various backgrounds, including real scenes and posters. Observations indicate that our method surpasses existing methods in word accuracy. Furthermore, our approach also excels in eliminating visual inconsistencies within the generated text region. In the fourth column, we compare the generation performance of various methods specifically for small-sized text. It can be observed that only our method is capable of generating tiny text effectively. In addition, we compare our method with the state-of-the-art (SOTA) direct generation approaches, as illustrated in Figure~\ref{fig:sota}. The results verify that our method effectively avoids the occurrence of irrelevant text. Finally, Figure~\ref{fig:small} demonstrates our remarkable capability in generating extremely small-sized text. In the Appendix~\cite{li}, a detailed exhibition of the text image customization and editing capabilities is provided, accompanied by additional comparative figures with other direct generation methods.

\subsection{Experiments in Boosting Scene Text Detectors}
\label{sec:Scene Text Detectors}
One of the most crucial downstream tasks achieved by our method is the synthesis of a scene text dataset with accurate annotations, aimed at enhancing existing text detectors. To this end, we have generated a synthetic dataset named SynthGlyph 10K.
     
We choose previous visual text blending methods which aim to generate synthetic data as fair comparison, including SynthText, VISD, SynthText3D, and UnrealText in some scene text detection benchmarks like ICDAR2013, ICDAR2015, and TotalText~\cite{totaltext}.

\paragraph{Pretraining} In this setup, we pretrain the DBNet using synthetic data and then conduct evaluations on real datasets. The results of this experiment are detailed in Table~\ref{tab_pretrain_db}. It is observed that the text detector achieves the best performance across all test sets when pretrained on our generated dataset. This success is attributed to the greater diversity of backgrounds in our synthetic data, which closely resemble real images, coupled with the advanced text rendering capabilities of GlyphOnly. Notably, our data demonstrates marked superiority in more challenging benchmarks like IC15 and TotalText, largely due to GlyphOnly's proficiency in small-scale text rendering.

\paragraph{Fine-tuning} We select the challenging IC15 dataset for our fine-tuning experiment (refer to Table~\ref{tab_ic15_db}), where the pre-trained detector is fine-tuned using real data. The results clearly show that the model trained on SynthGlyph outperforms others significantly. We are confident that our method can empower the development of large-scale scene text detectors by providing substantial training datasets.

\begin{table}[ht]
\vspace{-3pt}
\begin{center}
    
\caption{Scene text detection results of fine-tuning on IC15. DBNet model is pretrained on one of the synthetic datasets, fine-tuned on IC15, and evaluated on IC15's test dataset.}
\vspace{+12pt}
\begin{tabular}{@{}lccc@{}}
\toprule
Training Data & Precision & Recall & Hmean\\
\midrule
IC15 & 83.03 & 75.64 & 79.16 \\
IC15 + SynthText 10K & 88.79 & 80.50 & 84.44 \\
IC15 + VISD 10K & 90.10 & 81.08 & 85.35 \\
IC15 + SynthText3D 10K & 89.45 & 80.45 & 84.71 \\
IC15 + UnrealText 10K & 86.67 & 81.70 & 84.11 \\
IC15 + SynthGlyph 10K & 88.95 & 83.29 & \textbf{86.03}\\
\bottomrule
\end{tabular}

\label{tab_ic15_db}
\end{center}
\vspace{-5pt}
\end{table}

\subsection{Ablation Study}\label{sec:ablation}

\begin{table*}[ht]

\caption{Scene text detection results with different backgrounds. $\dag$ denotes that the generated background is utilized. ‘full’ implies the utilization of the whole background dataset; otherwise, a set of 5K backgrounds is employed. We use SynthText to generate datasets specifically for pretraining DBNet and evaluate its performance on real-world datasets.
}
\vspace{+10pt}
\centering
\begin{tabular}{@{}cccccccc@{}}
\toprule
\multirow{2}{*}{Training Data}&\multirow{2}{*}{Background}& \multicolumn{3}{c}{IC13} & \multicolumn{3}{c}{IC15}\\
\cmidrule(l){3-5} \cmidrule(l){6-8}
& & Precision & Recall & Hmean & Precision & Recall & Hmean\\
\midrule
SynthText 10K & Original 5K & 74.52 & 56.35 & 64.17 & 52.26 & 46.85 & 49.40 \\
SynthText (full) 10K & Original 10K & 75.24 & 63.56 & 68.91 & 70.27 & 46.89 & 56.25 \\
$\text{SynthText}^\dag$ 10K & Our 5K &75.03 & 65.02 & 69.67 & 69.38 & 48.77 & 57.28\\
$\text{SynthText}^\dag$ (full) 10K & Our 10K & 76.47 & 67.67 & \textbf{71.80} & 71.35 & 49.40 & \textbf{58.38} \\
\bottomrule
\end{tabular}
\label{tab_bg}
\end{table*}

\begin{table}[ht]
\vspace{-5pt}
\caption{Ablation study results for the recognition loss and the usage of text encoder and glyph encoder.}
\vspace{+10pt}
\centering
\begin{tabular}{@{}ccccc@{}}
\toprule
Glyph & Text & $\lambda _{rec}$ & Accuracy & 1-NED \\ \midrule
\usym{2713} & & 0 & 65.79 & 71.87 \\
\usym{2713} & & 0.01 & 66.88 & 72.58 \\
\usym{2713} & & 0.03 & \textbf{66.99} & \textbf{72.75} \\ 
\usym{2713} & & 0.1 & 66.41 & 72.54\\ 
& \usym{2713} & 0.03 & 63.08 & 70.60\\
\usym{2713} & \usym{2713} & 0.03 & 65.53 & 71.79 \\
\bottomrule
\end{tabular}
\label{tab_reg_loss}
\vspace{-5pt}
\end{table}

\begin{table}[ht]

\caption{Comparison between the accuracy of generated text with and without the addition of the background condition.}
\vspace{+10pt}
\centering
\begin{tabular}{@{}ccc@{}}
\toprule
Background condition & Accuracy & 1-NED \\ \midrule
 & 67.29 & 72.93 \\ 
\usym{2713} & 66.99 ($\downarrow$0.3) & 72.75 ($\downarrow$0.18) \\
\bottomrule
\end{tabular}
\vspace{-5pt}
\label{tab_bg_a}
\end{table}

\paragraph{The Significance of Creating Backgrounds} To validate the effectiveness of our background generator, we produce 5K and 10K images to serve as the source backgrounds for synthesizing 10K datasets using the SynthText method \cite{SynthText}. According to the results shown in Table~\ref{tab_bg}, the detector trained with 5K backgrounds generated by our method outperforms those trained with the standard 10K fixed data from SynthText. Furthermore, the performance of the detector improves as the volume of our synthetic data increases. This finding suggests that our backgrounds match the distribution of real-world scenes more closely. Moreover, it proves the quality of the background we generated. The enhancement of the experimental results is closely associated with the detection and erasure process.

\paragraph{The Effectiveness of Glyph Condition} We replace the condition from glyph to text prompt, or retain both like \cite{glyphdraw} while keeping the remaining modules unchanged. The results, presented in Table~\ref{tab_reg_loss}, reveal that using only the glyph condition yields higher text generation accuracy. A combination of glyph and the text prompt results in a minor accuracy decrease, whereas relying solely on the text prompt leads to a substantial accuracy reduction. We contend that this discrepancy is primarily due to the domain gap between natural language and visual texts.

\paragraph{The Weight of Recognition Loss} The experimental findings are showcased in Table~\ref{tab_reg_loss}, where a range of $\lambda_{rec}$ values ([0, 0.03, 0.01, 0.1]) are evaluated to examine their impact on the results. It is evident that the highest accuracy is achieved when $\lambda_{rec}$ is set to 0.03.

\paragraph{The Effectiveness of Background Condition} Initially, we conduct qualitative experiments to demonstrate that our background condition effectively complements surrounding information. As observed in Figure~\ref{fig:bg_ablation}, the inclusion of the background condition notably reduces fusion artifacts (refer to the first and second columns). Furthermore, it enables more natural blending with local regions, even in complex scenarios such as object occlusion (see the third column). Additionally, style deviations in the blending regions are significantly mitigated (refer to the last column). Our results, detailed in Table~\ref{tab_bg_a}, also confirm that the introduction of the background condition does not impede accurate text rendering, with only a slight decrease in accuracy.

\section{Conclusion}

In this paper, we revisit the process of visual text blending and introduce a novel two-stage approach. Separating background and text generation in scene text images addresses limitations in direct text-to-image methods, allowing better control over text elements and the background. Our method includes the development of a background generator to synthesize high-fidelity and text-free backgrounds. Additionally, we present GlyphOnly - a diffusion-based model specifically designed to render texts with high accuracy and visual consistency. GlyphOnly is particularly effective in addressing the challenges of generating small-scale texts. Utilizing our proposed method, we delve into several downstream applications, notably in the synthesis of scene text datasets. Our synthesized data greatly enhances the performance of existing scene text detectors. However, the two-stage method results in slower speeds (30-40s per image). Future research should focus on optimizing speed and extending the method to video text generation. Considering that text is fine-grained, this inspires exploring fine-grained object generation and line refinement.

\begin{ack}
This work is supported by the National Natural Science Foundation of China (Grant NO 62376266), and by the Key Research Program of Frontier Sciences, CAS (Grant NO ZDBS-LY-7024).
\end{ack}
%%%%%%%%%%%%%%%%%%%%%%%%%%%%%%%%%%%%%%%%%%%%%%%%%%%%%%%%%%%%%%%%%%%%%%%%

%%% Use this environment to include acknowledgements (optional).
%%% This will be omitted in doubleblind mode.

%%%%%%%%%%%%%%%%%%%%%%%%%%%%%%%%%%%%%%%%%%%%%%%%%%%%%%%%%%%%%%%%%%%%%%%%

%%% Use this command to include your bibliography file.

\bibliography{main}

\end{document}